%% file: main.tex
\documentclass[10pt,twocolumn,letterpaper]{article}

\usepackage{wacv}
\usepackage{times}
\usepackage{epsfig}
\usepackage{graphicx}
\usepackage{amsmath}
\usepackage{amssymb}

\usepackage{caption}
\usepackage[font=small]{caption}
\usepackage{color}
\usepackage{subfigure}
\usepackage{epstopdf}
\usepackage{array}
\newcolumntype{P}[1]{>{\centering\arraybackslash}p{#1}}

\newcommand{\old}[1]{}

\usepackage[breaklinks=true,bookmarks=false]{hyperref}

\wacvfinalcopy 

\def\wacvPaperID{311} 

\begin{document}

\title{Good Similar Patches for Image Denoising}

\author{Si Lu\\
Portland State University\\
{\tt\small lusi@pdx.edu}
}

\maketitle

\begin{abstract}
   Patch-based denoising algorithms like BM3D have achieved outstanding performance. An important idea for the success of these methods is to exploit the recurrence of similar patches in an input image to estimate the underlying image structures. However, in these algorithms, the similar patches used for denoising are obtained via Nearest Neighbour Search (NNS) and are sometimes not optimal. First, due to the existence of noise, NNS can select similar patches with similar noise patterns to the reference patch. Second, the unreliable noisy pixels in digital images can bring a bias to the patch searching process and result in a loss of color fidelity in the final denoising result. We observe that given a set of good similar patches, their distribution is not necessarily centered at the noisy reference patch and can be approximated by a Gaussian component. Based on this observation, we present a patch searching method that clusters similar patch candidates into patch groups using Gaussian Mixture Model-based clustering, and selects the patch group that contains the reference patch as the final patches for denoising. We also use an unreliable pixel estimation algorithm to pre-process the input noisy images to further improve the patch searching. Our experiments show that our approach can better capture the underlying patch structures and can consistently enable the state-of-the-art patch-based denoising algorithms, such as BM3D, LPCA and PLOW, to better denoise images by providing them with patches found by our approach while without modifying these algorithms.
\end{abstract}
\vspace{-0.20in}

\vspace{-0.10in}
\section{Introduction}
\label{sec:intro}
\input{intro}

\vspace{-0.05in}
\section{Related Work}
\vspace{-0.05in}

\label{sec:related}
\input{related}

\section{Good Similar Patch Searching}

\label{sec:method}

\input{method}

\section{Experimental Results}
\label{sec:exp}
\input{exp}

\vspace{-0.10in}
\section{Conclusion}

\label{sec:concl}

This paper presented an approach to similar patch searching for patch-based denoising algorithms. By combing an unreliable pixel estimation algorithm and a clustering-based patch searching with adaptively estimated cluster numbers, better similar patches can be obtained for image denoising. We showed that similar patches collected by our approach can capture the underlying image intensity fidelity and the image structures more adequately. Experimental results demonstrated that our approach can effectively improve the quality of searched similar patches and can consistently enable better denoising performance for several recent patch-based methods, such as BM3D, PLOW and LPCA. Since patch-based methods have attracted a significant amount of research effort and provide the state-of-the-art performance, our method can be incorporated into all the patch-based denoising methods.

{\small
\bibliographystyle{ieee}
\bibliography{egbib}
}

\end{document}

%% file: intro.tex

Image capturing has become a daily practice these days and millions of images are taken every day. These images, however, sometimes suffer from noise due to sensor errors. To solve this problem, image denoising has been extensively studied. Numerous image denoising methods~\cite{chen:pg,bm3d,ksvd-denoise,ddid,mildenhall2017burst,Roth:foe,bilateral,Wang_2017_ICCV,Xu_2018_ECCV,Xu_2017_ICCV,epll} have been developed. Recently, patch-based approaches~\cite{nlm,cluster-diction,plow,chen:pg,dong2013nonlocally,wnnm,mairal2009non,xu:pgpd,lpg-pca} have shown great success. Their key idea is to exploit the recurrence of similar patches in a noisy input image to estimate the underlying patch structure. The quality of the selected similar patches is therefore an important factor that can influence the final denoising result. 

\begin{figure} [tb]
    
    \begin{center}
      \begin{tabular}{cc}
        \hspace{-0.09in}\includegraphics[width=0.24\textwidth]{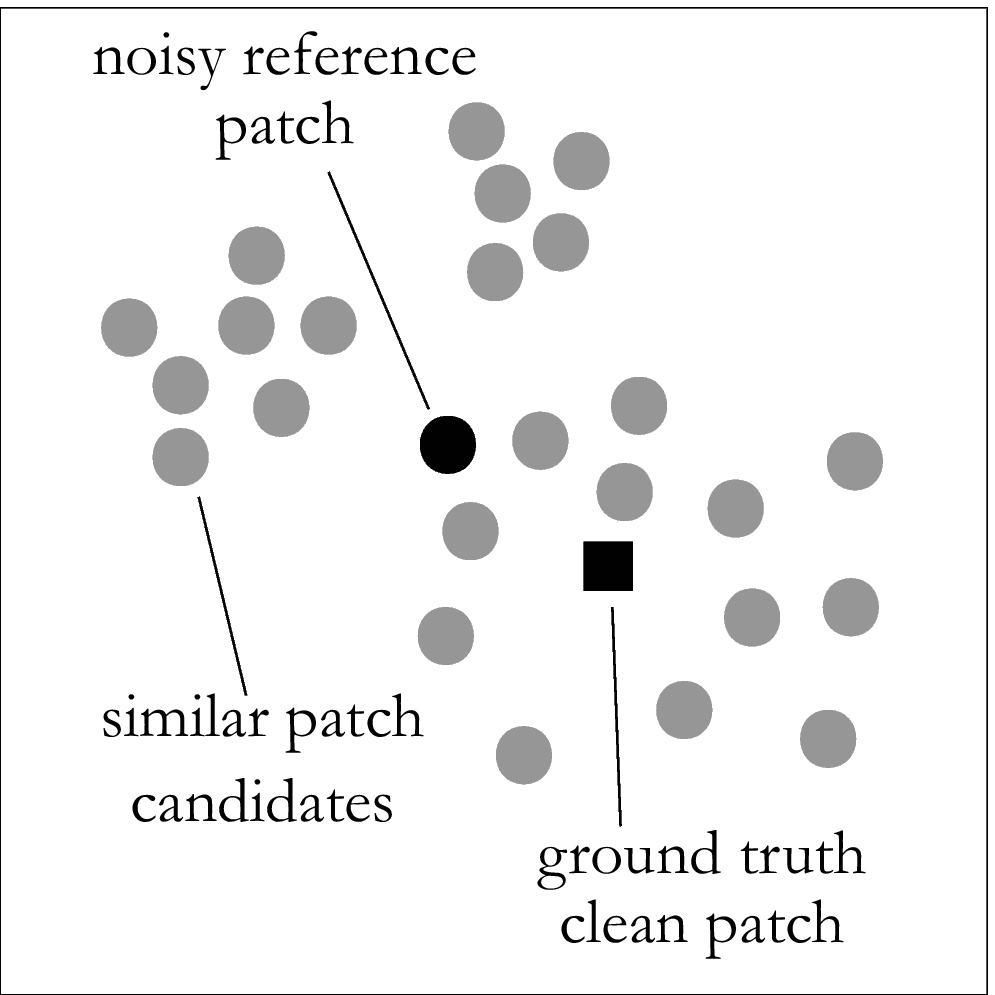}  &
		\hspace{-0.165in}\includegraphics[width=0.24\textwidth]{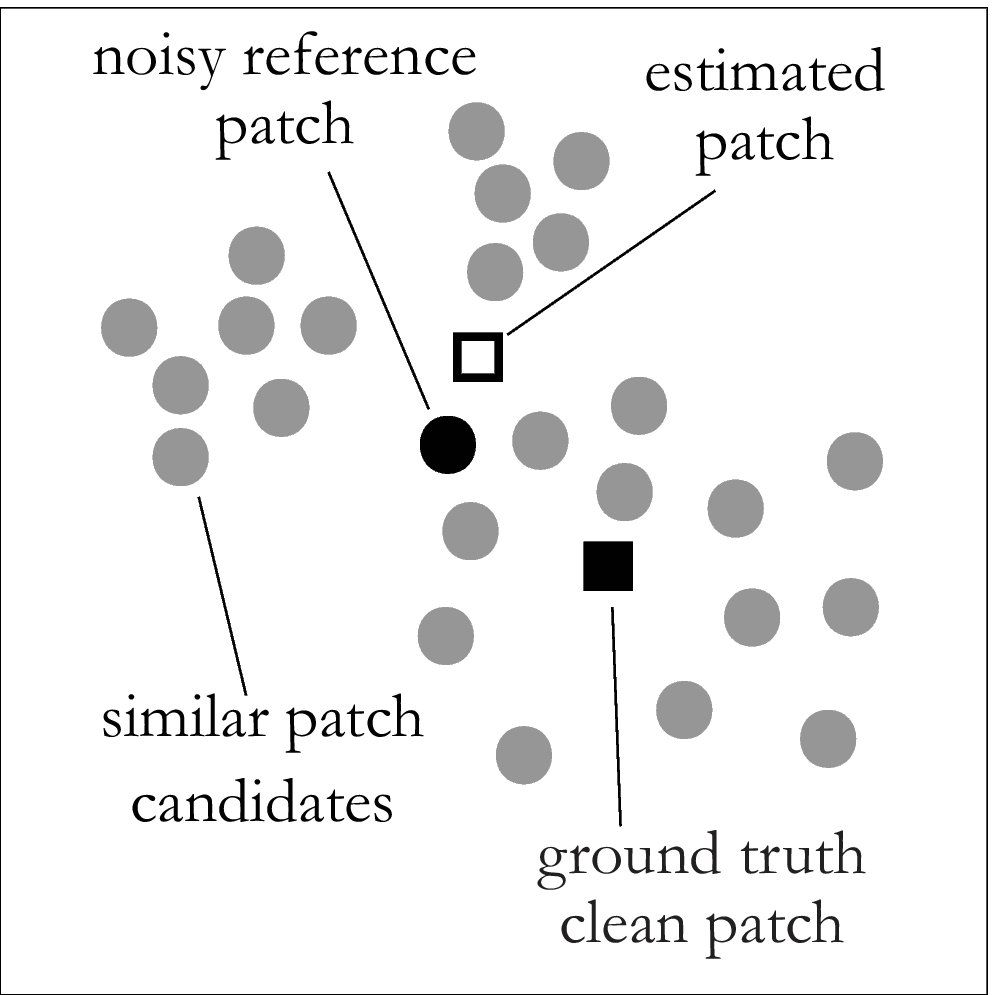}
      \end{tabular}
	  \footnotesize{
      \caption{ Nearest Neighbour Search (NNS) is not optimal for patch searching. Given a reference patch and a set of similar patches obtained by NNS (left), the estimated patch is close to the noisy reference patch rather than the clean ground truth patch(right). In this figure, patches are represented as 2D feature points for the convenience of visualization.
      \label{fig:motiv_nnslimit} }
      }
    \end{center}
	\vspace{-0.25in}
\end{figure}

\begin{figure*}[htb]
    \begin{center}
      \begin{tabular}{cccc}
		\hspace{-0.08in}\includegraphics[width=.245\textwidth]{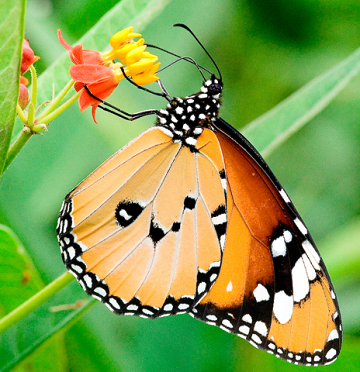}          &
        \hspace{-0.12in}\includegraphics[width=.245\textwidth]{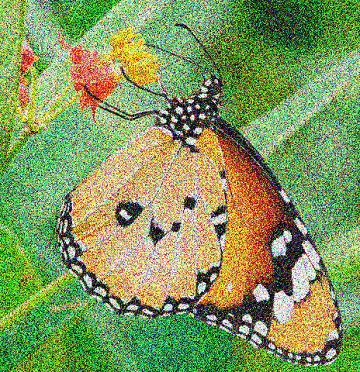}       &
		\hspace{-0.12in}\includegraphics[width=.245\textwidth]{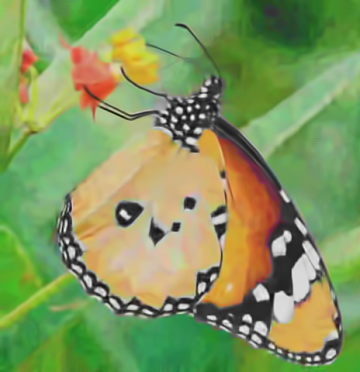}  &
		\hspace{-0.12in}\includegraphics[width=.245\textwidth]{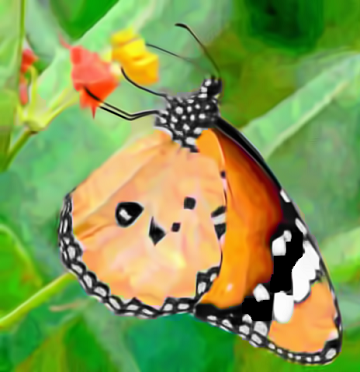} \\ 		
		Ground truth &Noisy image& BM3D, PSNR=18.97& BM3D$_{bst}$, PSNR=22.18    
      \end{tabular}
	  \vspace{-0.03in}
	  \begin{tabular}{c}
	    (a) Visual comparison on image \emph{Monarch$_c$} between BM3D and BM3D$_{bst}$.
	  \end{tabular}
	  \begin{tabular}{cccc}
		\hspace{-0.08in}\includegraphics[width=.245\textwidth]{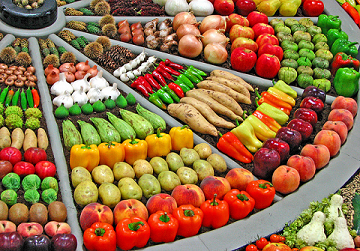}               &
		\hspace{-0.12in}\includegraphics[width=.245\textwidth]{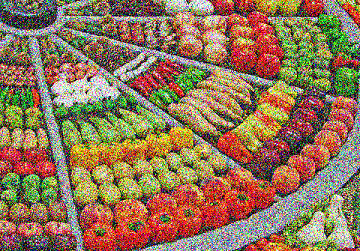}           &
		\hspace{-0.12in}\includegraphics[width=.245\textwidth]{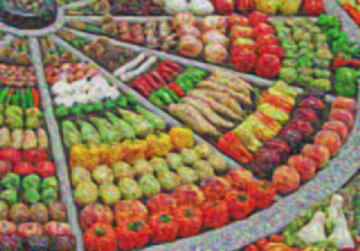}      &
		\hspace{-0.12in}\includegraphics[width=.245\textwidth]{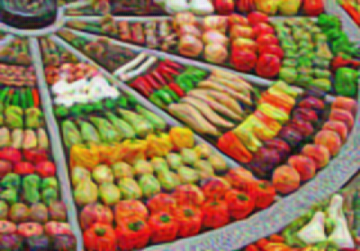}\\		
		Ground truth &Noisy image& PLOW, PSNR=17.20& PLOW$_{bst}$, PSNR=17.97 
      \end{tabular}
	  \vspace{-0.03in}
	  \begin{tabular}{c}
	    (b) Visual comparison on image \emph{Vegetables$_c$} between PLOW and PLOW$_{bst}$.
	  \end{tabular}
	  \begin{tabular}{cccc}
		\hspace{-0.08in}\includegraphics[width=.245\textwidth]{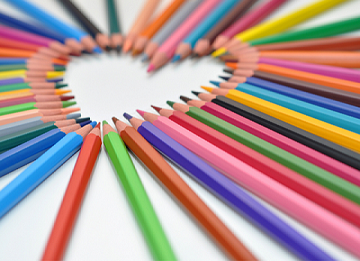}              &
		\hspace{-0.12in}\includegraphics[width=.245\textwidth]{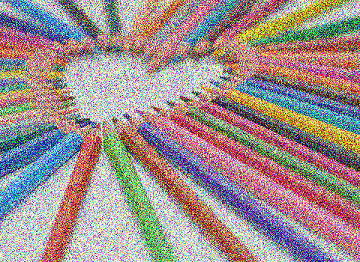}          &
		\hspace{-0.12in}\includegraphics[width=.245\textwidth]{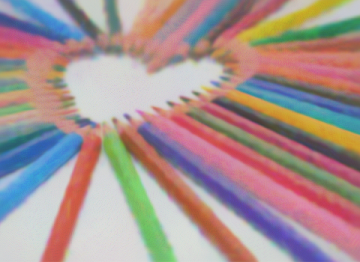}      &
		\hspace{-0.12in}\includegraphics[width=.245\textwidth]{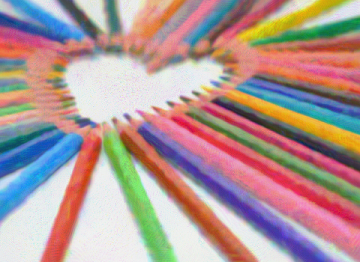} \\	
		Ground truth &Noisy image& LPCA, PSNR=20.21& LPCA$_{bst}$, PSNR=24.22 		
      \end{tabular}
	  \begin{tabular}{c}
	    (c) Visual comparison on image \emph{Pencils$_c$} between LPCA and LPCA$_{bst}$.
	  \end{tabular}
	  \vspace{-0.07in}
      \caption{Visual comparison on denoised images by the original algorithms and our boosted algorithms. \label{fig:visual_01}}
    \end{center}
	\vspace{-0.25in}
\end{figure*}

Nearest Neighbour Search (NNS), which selects each patch's nearest neighbours as potential similar patches, is widely used for patch searching due to its simplicity. However, due to the existence of noise, this method can introduce bias to the search results of similar patches. As shown in Figure~\ref{fig:motiv_nnslimit} (a), the reference patch is corrupted by noise (marked in red). NNS thus prefers similar patches that contain the same noise pattern as the reference one. Consequently, the estimated patch is close to the noisy reference rather than the ground truth clean patch (marked in green). This bias can finally retain the noise pattern in the denoised image, as shown in Figure~\ref{fig:motiv_nnslimit} (b).


In this paper, we present a patch searching method to find a set of good similar patches for patch-based denoising algorithms, such as BM3D~\cite{bm3d}, LPCA~\cite{lpg-pca} and PLOW~\cite{plow}. We consider that \emph{a set of good similar patches should be as similar to the noise-free version of the reference patch as possible rather than the noisy reference patch}. Our assumption is that the distribution of these good similar patches can be approximated as a Gaussian function although this distribution is not necessarily centered around the noisy reference patch. This is a popular assumption in image denoising. Based on this assumption, we develop the following patch searching method. We first use Nearest Neighbour Search to obtain a set of candidate similar patches for each reference patch. We then model the distribution of these candidate patches as a mixture of Gaussian components and cluster them into several sub-groups. We finally take the sub-group that contains the reference patch as the set of similar patches for denoising. To further improve the quality of similar patches, we pre-process the input noisy images using an unreliable pixel estimation to eliminate the influence of unreliable pixels.


This paper contributes to the problem of image denoising with a way to find better similar patches for image denoising. This paper shows that the performance of existing patch-based denoising algorithms can be consistently improved by inputting them with a better set of similar patches. Our method has an advantage that no modification needs to be made to these existing denoising algorithms. As shown in Figure~\ref{fig:visual_01}, our method can enable the state-of-the-art denoising methods like BM3D, LPCA and PLOW to better denoise noisy images. We expect that our method can also be married to other patch-based denoising methods.

%% file: related.tex
Image denoising aims to recover the underlying clean image $I$ from the corrupted noisy observation $I_n$, which can be modeled as $I_n = I+N$, where $N$ is the additive noise. The additive noise is often assumed to have a Gaussian distribution with zero mean or a Poisson distribution~\cite{lowlightpos02,lowlightpos} and thus is usually denoted as White Gaussian noise or Poisson noise. A variety of denoising methods~\cite{natural-denoising,Patch-comp-denoising,bilateral} have been developed based on different assumptions, models, priors or constraints.

As image denoising is an ill-posed problem, many methods exploit priors to denoise images~\cite{MLP-denoising,denoising-bounds}. The EPLL algorithm~\cite{epll} learns Gaussian Mixture Models from external clean patches and iteratively restores the underlying clean image via Expected Patch Log Likelihood (EPLL) maximization. Yue et al.~\cite{cidweb} retrieved similar web images to help with image restoration by combining 3D block denoising in both the spatial and frequency domains. Zontak et al.~\cite{internal-stat} analysed the internal statistics of natural images and exploited internal patch recurrence across scales to restore the underlying image content from the noisy input~\cite{across-scale-denoise}. Mosseri et al.~\cite{inter-extern-comb} proposed a combining method to locally select external or internal priors for image denoising. The success of KSVD~\cite{ksvd-basic,ksvd-denoise} brought a new trend of image denoising by training dictionaries for patch restoration via sparse representation, including a number of variations and extensions~\cite{cluster-diction,diction-02}.


Based on the assumption that the underlying similar patches lie in a low-dimensional subspace, low-rank constraint-based methods~\cite{dong2013:lowrank,lu2014depth,high-ord-svd} have shown promising performance on image denoising in recent years. Gu et al.~\cite{wnnm} considered the patch denoising as a low-rank matrix factorization problem for similar patches and proposed a Weighted Nuclear Norm Minimization (WNNM) process to assign different weights for different singular values.

Similar patch-based methods~\cite{nlm,cluster-diction,plow,chen:pg,dong2013nonlocally,wnnm,mairal2009non,xu:pgpd,lpg-pca} are among the most popular denoising techniques and have shown great success on image denoising. All these methods exploit the image non-local self-similarity prior\textemdash natural image patterns repetitively occur across the whole image. The recent benchmark BM3D algorithm~\cite{bm3d} applies collaborative filtering in the transform domain on 3D similar patch blocks to estimate the latent image structures. LPCA~\cite{lpg-pca} performs PCA on similar patch groups and PLOW~\cite{plow} restores the latent clean image structures using similar patches via an adaptive Wiener filter. However, as demonstrated in Section~\ref{sec:intro}, Nearest Neighbour Search is not optimal for similar patch searching, especially in images with heavy noise. Our approach aims to solve this problem via a clustering-based patch searching approach. Our method modifies both the similar patch locations as well as the input similar patch values for similar patch-based denoising algorithms.


While most patch-based denoising techniques use Nearest Neighbour Search, clustering has already been proved effective for similar patch searching. Chatterjee et al.~\cite{plow} used LARK~\cite{lark} features
to cluster the noisy image into regions with similar patch patterns. However, in images with heavy noise, LARK feature fails to capture properties of the underlying patches and thus can not provide accurate clustering results. Chen et al.~\cite{chen:external} collected external clean patches to train a fixed set of Gaussian Mixture Models and cluster all noisy patches into clusters defined by the models. However, this method independently assigns each patch to a cluster with the maximum likelihood while our method exploits the overall similar patch distributions for clustering. Our method also differs from this method in that 1) our method does not require external clean patches for training, 2) rather than using fixed Gaussian Mixture Models for all input noisy images, our method tries to estimate dedicated Gaussian Mixture Models for each individual similar patch group and thus works better across images with different contents and properties, and 3) our method adaptively estimates the number of clusters via an optimization process while the previous methods used fixed cluster numbers. Thus, our method better exploits individual patch coherence and can recover patch structure details more adequately. 

Lotan and Irani provided needle-match~\cite{needleMatch}, an effective patch descriptor for patch searching for applications like denoising. While we share the same goal, our work is actually orthogonal to theirs. They find nearest neighbors as good patches according to their descriptor instead of pixel values while our method finds good patches by selecting a subset from the set of patches, which are currently found using nearest neighbor search based on pixel colors. We expect that combining our searching algorithm together with their powerful descriptor will further help denoising.

%% file: method.tex
\begin{figure*} [tb]
    \vspace{-0.05in}
    \begin{center}
      \begin{tabular}{c}
        \hspace{-0.08in}\includegraphics[width=1.0\textwidth]{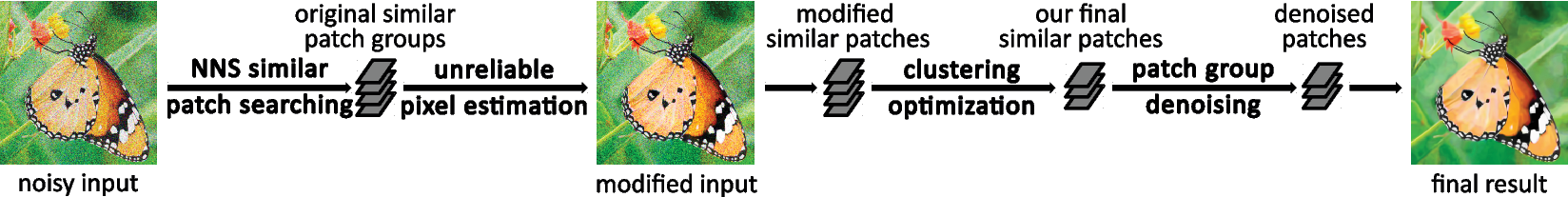}  \\
      \end{tabular}
	  \vspace{-0.05in}
	  \footnotesize{
      \caption{ Algorithm overview. We embed our similar patch searching approach by first inserting an unreliable pixel estimation (section~\ref{sec:unreliable_est}) given the original similar patches obtained by NNS. The modified patches (with only values possibly changed) are used as clustering candidates. Patches obtained by the clustering optimization (section~\ref{sec:clustering_opt}) are then fed to the original denoising procedures. Note that NNS is only applied once and is not re-applied to the modified input image.
      \label{fig:overview}}}
    \end{center}
	\vspace{-0.2in}
\end{figure*}

Most existing patch-based image denoising methods share a common two-step pipeline: first find a set of similar patches for each reference patch and then perform patch group denoising to obtain the denoising result for the reference patch. Our similar patch searching algorithm can be married with a patch-based denoising method by replacing its original similar patch searching algorithm with ours or embedded into the denoising method in-between these two steps, as illustrated in Figure~\ref{fig:overview}. The overall goal of our algorithm is to provide a set of good similar patches to enable better image denoising.

We consider good similar patches for a reference patch should be close to its noise-free version. The widely used Nearest Neighbour Search (NNS) finds a set of similar patches that are closest to the reference. As shown in Figure~\ref{fig:insight} (a), many of these patches are faraway from the noise-free reference patch (indicated in green). If such a set of patches are used for denoising, the denoised result (indicated in blue) can deviate from the noise-free patch significantly, as illustrated in Figure~\ref{fig:insight} (b). Our assumption is that the distribution of these good similar patches can be modeled as a Gaussian function approximately around the noise-free reference patch, as illustrated in Figure~\ref{fig:insight} (d). This is a popular assumption in image denoising. Based on this assumption, we can formulate the problem of good similar patch searching as a Gaussian Mixture Model (GMM)-based clustering and group candidate patches into several clusters. The patches in the cluster containing the reference patch are selected as good similar patches for denoising.

Our algorithm starts by using NNS to find a set of candidate similar patches for each reference patch. While NNS cannot find an optimal set of similar patches, it can filter out most outliers. We then select a set of good similar patches from these candidates. In the rest of this section, we first describe our GMM clustering-based good similar patch searching algorithm, then discuss how to detect unreliable pixels and update them to further improve our method, and finally discuss the complexity of our method.

\subsection{Similar Patch Clustering}
\label{sec:clustering_opt}
\vspace{-0.05in}

\begin{figure} [b]
\vspace{-0.15in}
    \begin{center}
      \begin{tabular}{cccc}
        \hspace{-0.08in}\includegraphics[width=0.117\textwidth]{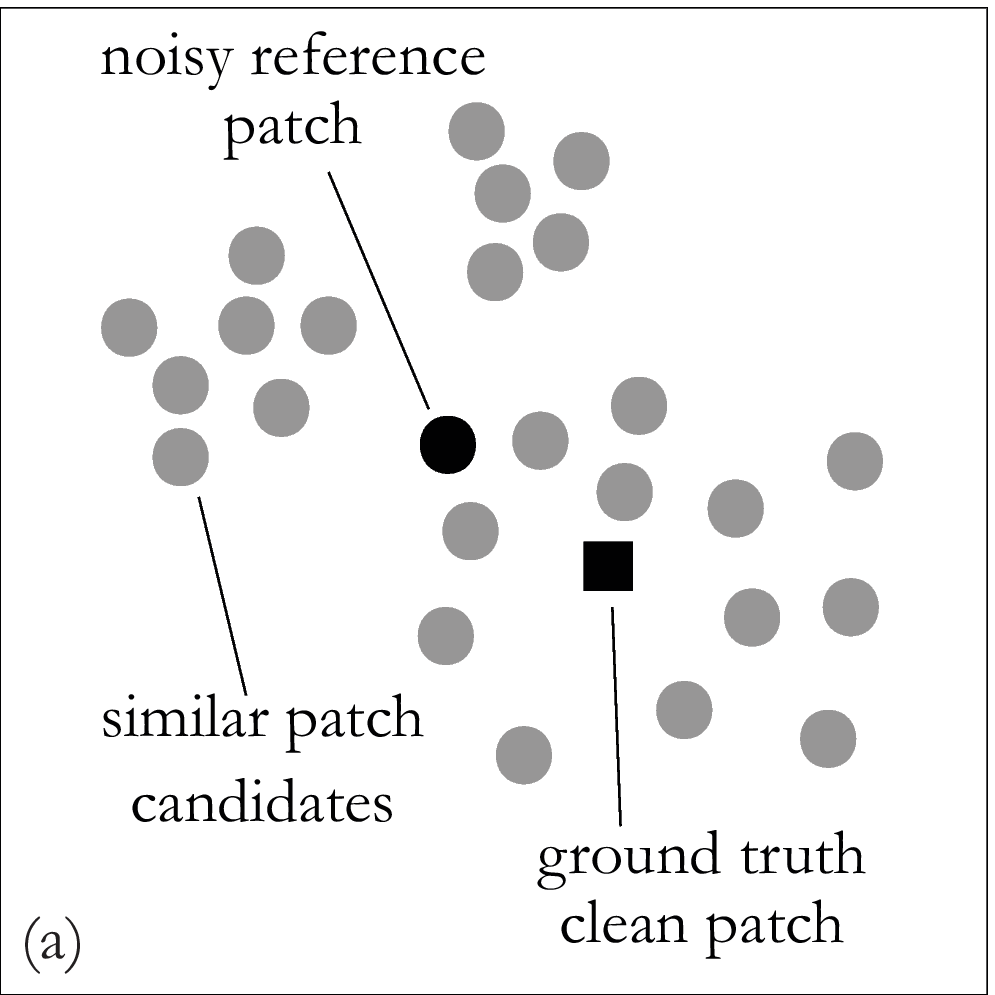} &
		\hspace{-0.15in}\includegraphics[width=0.117\textwidth]{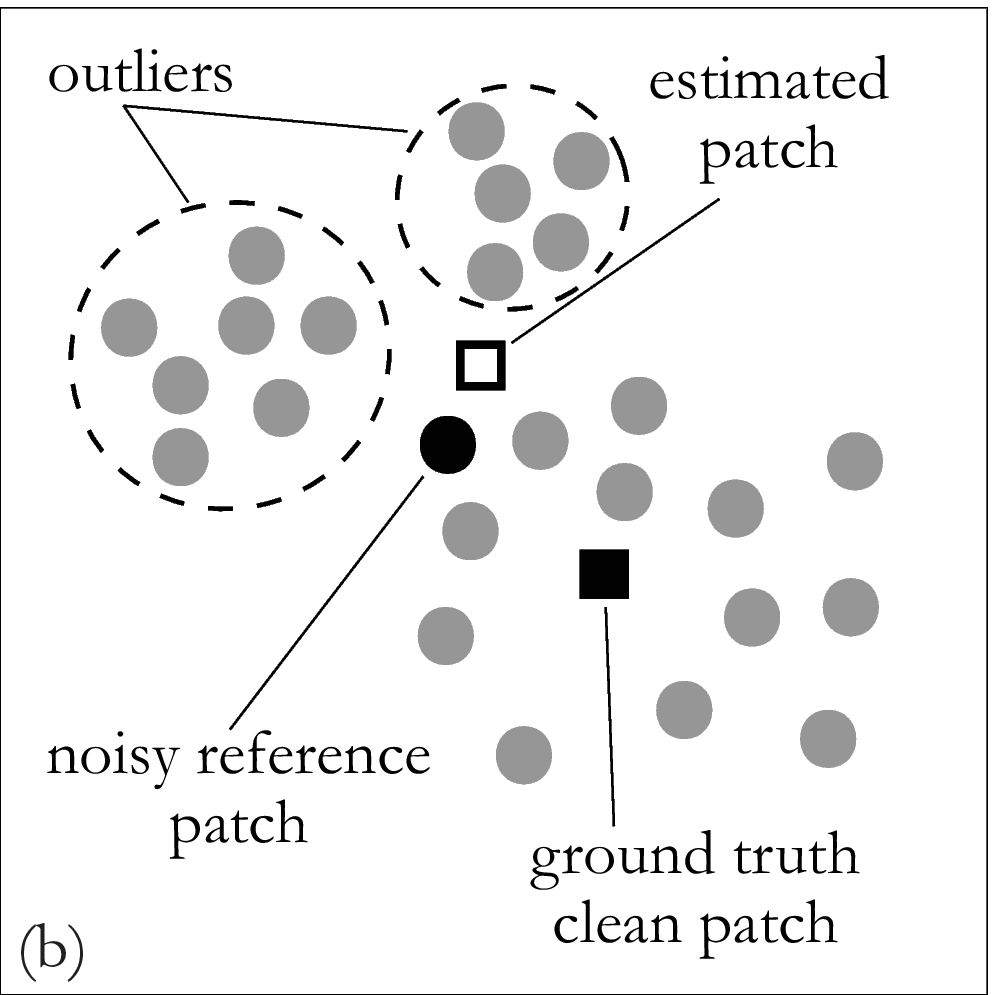} &
		\hspace{-0.15in}\includegraphics[width=0.117\textwidth]{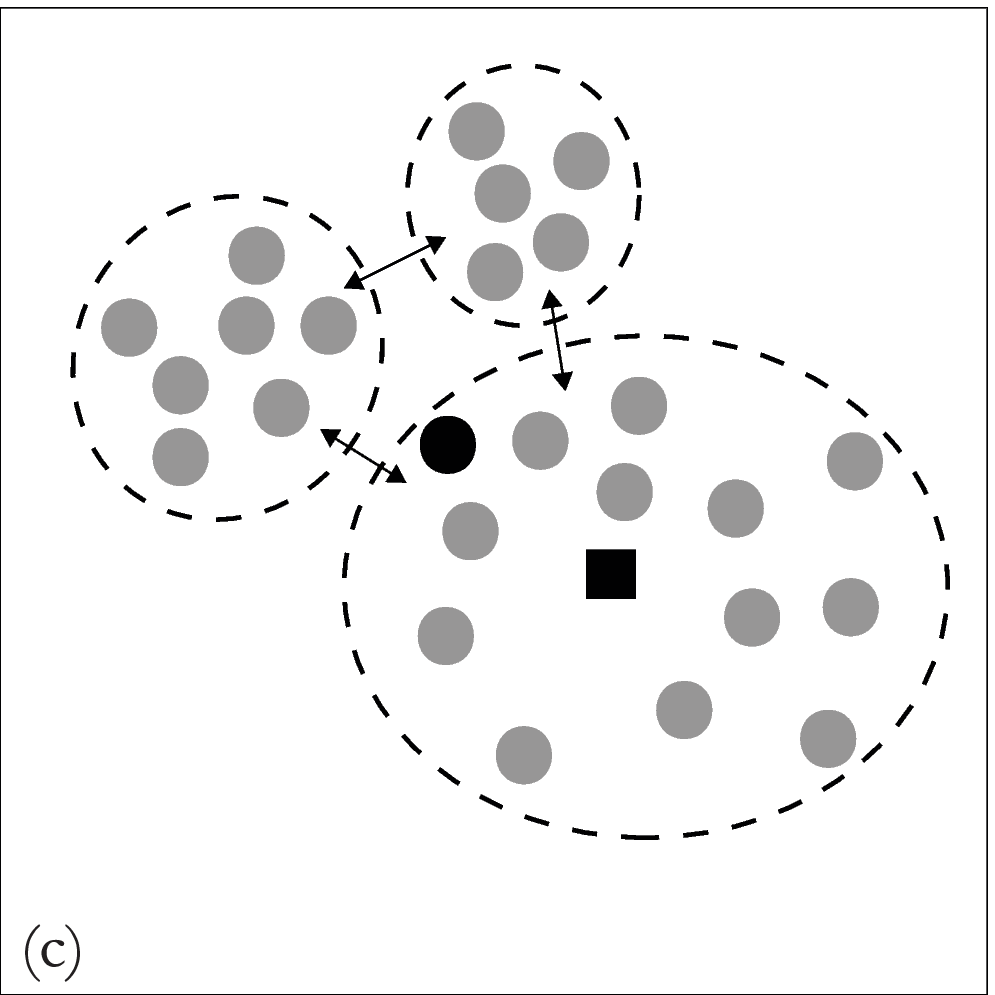} &
		\hspace{-0.15in}\includegraphics[width=0.117\textwidth]{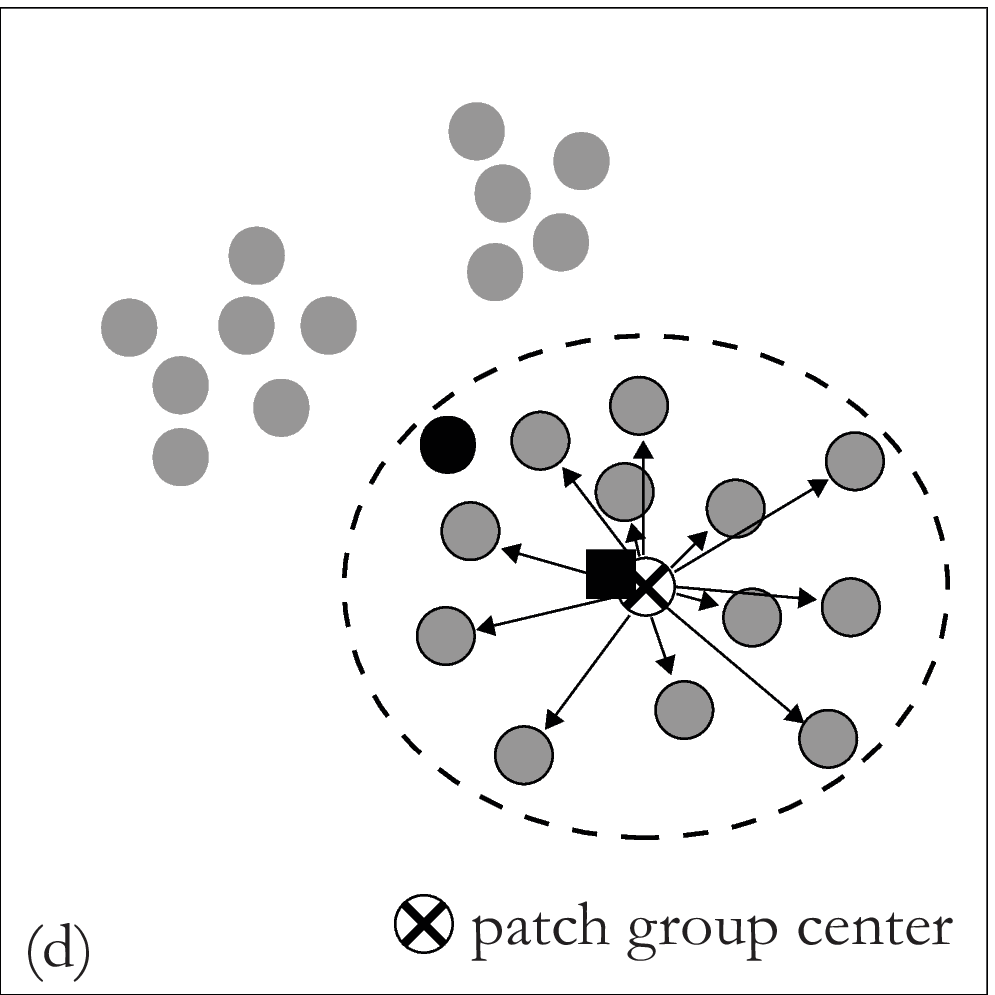} 
      \end{tabular}
	  \vspace{-0.05in}
      \caption{ \footnotesize{ Insight of our patch searching approach. (a) A noisy reference patch and its NNS candidates. (b) NNS only considers the patch distance to the noisy reference and brings bias to the estimated patch. A good patch searching approach should (c) be able to distinguish valid similar patches and outliers by classifying them into different sub-groups, (d) select similar patches that are closest to the patch group center rather than the noisy reference.}
      \label{fig:insight}}
    \end{center}
	\vspace{-0.15in}
\end{figure}

As shown in Figure~\ref{fig:insight} (a) and (b), NNS only considers the distance to the noisy reference in selecting similar patches. Outliers that are close to the noisy reference rather than the ground truth clean patch are thus selected as valid similar patch candidates. Ideally, the overall patch distribution should be carefully investigated to find good similar patches. For the example in Figure~\ref{fig:insight} (c), all the candidate patches from NNS can be roughly clustered into three groups. For the group that contains the reference patch, its center is closer to the ground truth clean patch than to the noisy reference patch, as shown in Figure~\ref{fig:insight} (d).

We model each patch group as a Gaussian function and then the whole set of candidate patches can be approximated as a Gaussian Mixture Model (GMM). Specifically, given a set of $m$ $n$-dimensional candidate patches $Q=\{q_i|_{i=1,...,m}\}$, we aim to estimate a GMM with $K$ components. We denote $\theta=\{\pi_k, \mu_k, R_k\}_{k = {1...K}}$ as the parameters of the GMM, in which $R_k$ is the covariance matrix, $\mu_k$ is the centroid vector and $\pi_k$ is the probability that a given patch comes from the $k$th cluster. We regularize the estimation of the GMM using the Minimum Description Length (MDL) criteria introduced by Rissanen~\cite{mdl_01} as follows.

\vspace{-0.15in}
\begin{equation} \label{eq: mdl_1}
    MDL(K,\theta)=-\log p_{_Q}(Q|K,\theta)+\lambda L \log (mn),
\end{equation}
\vspace{-0.05in}

\noindent where the first term is a data term that encourages smaller intra-patch distance in each cluster as follows.

\vspace{-0.10in}
\begin{equation} \label{eq: mdl}
    -\log p_{_Q}(Q|K,\theta) = \sum_{k=1}^K{\sum_{j=1}^{m_k}||q_j-\mu_k||^2},
\end{equation}
\vspace{-0.10in}

\noindent where $m_k$ and $\mu_k$ are the number of patches in the $k$th cluster and its centroid. The second term is a regularization term that penalizes a large number of clusters. $\lambda$ is a parameter that balances the contribution of the data term and the regularization term. $L$ is the number of parameters required to define $\theta$ and is defined as

\vspace{-0.10in}
\begin{equation}
	L=K(1+n+(n+1)n/2)-1
\end{equation}
\vspace{-0.10in}

Based on an observation that candidate similar patches with complex structures are more likely to be generated from more clusters, we adaptively compute $\lambda$ according to the average gradient magnitude of the reference patch to prefer large cluster numbers for similar patches with complex structures.

\vspace{-0.10in}
\begin{equation} \label{eq: grad_lambda}
    \lambda=\alpha \exp(-\frac{1}{\beta}(\frac{1}{n}\sum_{i=1}^{n}||\nabla I_i||)^2)
\end{equation}
\vspace{-0.10in}

\noindent where the image gradient of each pixel $\nabla I_i$ can be computed from a preliminarily denoised version of the noisy input, since most patch-based denoising methods either generate a basic estimation for better patch searching or perform iterative image denoising. $\alpha$ and $\beta$ are two empirically selected parameters.

\begin{figure}[t]

\begin{center}
    \begin{tabular}{ccccc}
        \hspace{-0.10in}\includegraphics[width=.09\textwidth]{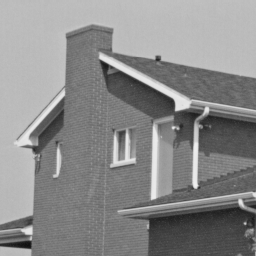} &
		\hspace{-0.12in}\includegraphics[width=.09\textwidth]{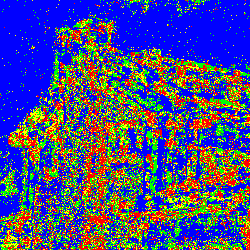}         &
		\hspace{-0.12in}\includegraphics[width=.09\textwidth]{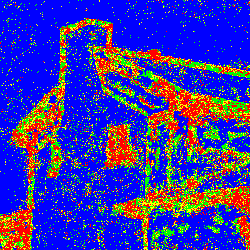}         &
		\hspace{-0.12in}\includegraphics[width=.09\textwidth]{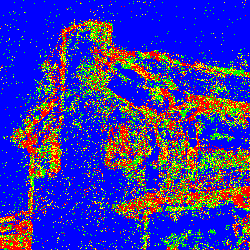}        &
		\hspace{-0.12in}\includegraphics[width=.09\textwidth]{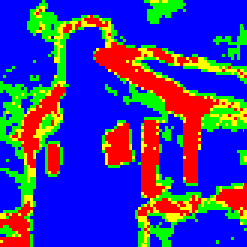} \\
		
	   \hspace{-0.10in}(a) clean        &
	   \hspace{-0.12in}(b) \footnotesize $\sigma=20$   &
	   \hspace{-0.12in}(c) \footnotesize $\sigma=60$   &
	   \hspace{-0.12in}(d) \footnotesize $\sigma=100$  &
	   \hspace{-0.12in}(e) \footnotesize our CN
	
    \end{tabular}	
	\vspace{-0.05in}	
    \caption{\footnotesize{Optimal cluster numbers (CN) at different noise levels. (a) The clean image. (b)-(d) Ground truth optimal cluster numbers for different noise levels ($\sigma=20,60,100$, respectively). (e) The cluster number map estimated by our approach for $\sigma=100$. In the cluster number maps (b), (c), (d) and (e), red color indicates larger cluster numbers.}\label{fig:optimal_CN}}
\end{center}
\vspace{-0.35in}
\end{figure}

To verify this observation, we select images with different noise levels and examine how gradients affect the optimal cluster numbers. Specifically, we select the optimal cluster number for a patch as the one that leads to the best denoising result according to the corresponding ground truth clean patch. We show the results in Figure~\ref{fig:optimal_CN} where the intensity value of each pixel in the cluster number maps (b), (c), and (d) encodes the optimal number of clusters for the patch centered at that pixel. This shows that the cluster number typically increases with the magnitude of the image gradient, especially at higher noise levels. We also show a cluster number map for $\sigma=100$ estimated by our clustering patch searching approach in Figure~\ref{fig:optimal_CN} (e). It shows that our optimization approach properly estimates the cluster numbers for patches that are close to the optimal ones.

\subsubsection{Optimization Reduction}
\label{sec:opti_reduct}

Minimization for Equation~\ref{eq: mdl_1} can be solved by a modified Expectation-Maximization (EM) algorithm proposed by Boumam \emph{et al.}~\cite{mdlcluster}. However, directly solving this optimization for the cluster number $K$ and the clustering results simultaneously is time-consuming. We reduce this problem with two ideas. First, we observe that the number of the clusters is relatively small and therefore use brute-force search to find an optimal cluster number since the searching space of $K$ is small. This step will not affect the optimality. Second, we use a fast K-means++~\cite{kmeanspp} algorithm to solve the GMM clustering problem given $K$. Note, given a cluster number, the regularization term is a constant and therefore can be ignored.

\subsection{Unreliable Pixel Detection and Update}
\label{sec:unreliable_est}


For a noisy pixel in a local reference patch, we can retrieve a group of similar pixels from the corresponding similar patches. These similar pixels can be used to estimate the latent clean intensity. However, outliers with pixel intensity values far away from the center might exist and the final estimated intensity would be biased.

We hence use a simple but effective unreliable pixel estimation (UPE) algorithm to detect outliers and accordingly modify their intensities in the input. For a reference pixel $x_r$ and its $m$ similar pixels $\{x_i|_{i=1,\cdots,m}\}$ ordered in intensities, we first compute two dynamic thresholds at both the high and low end ($t_{l}=max(0, x_M-\gamma\sigma)$ and $t_{h}=min (255, x_M+\gamma\sigma)$), where $x_M$ is the median and $\sigma$ is the standard deviation of the $m$ pixels. $\gamma$ is an empirically selected constant parameter with default value 2 for BM3D/PLOW and 4 for LPCA. Pixels with intensities beyond these two thresholds are then discarded. Suppose $n_l$ pixels have intensities smaller than $t_l$ and $n_h$ pixels have intensities larger than $t_h$, we set $t=\max(n_l,n_h)$ and discard $t$ pixels at both ends. A truncated mean $\hat{y}$ is then estimated by averaging the remaining pixels.




For ones with many outliers, the small amount of remained pixels may lead to inaccurate estimations. In this case, we find that the threshold $t_l$ or $t_h$ is a good approximation of $\hat{y}$ and directly assign $t_l$ or $t_h$ according to the number of outlier pixels at both the high and low end. For pixels with only a few outliers, we simply keep its intensity unchanged. Since each reference pixel $x_r$ is retrieved from a patch and most denoising algorithms use overlapped patches, each pixel can have multiple estimations. We simply use their weighted mean as the final modified intensity $y$ in the modified noisy input image, where the weight is set to $t$ to give more credits to pixels with more outliers.



\begin{figure} [b]
\vspace{-0.15in}
\begin{center}
\begin{tabular}{cccccc}
\footnotesize
    \hspace{-0.12in}\includegraphics[width=.068\textwidth]{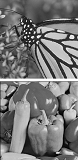}      &
    \hspace{-0.15in}\includegraphics[width=.068\textwidth]{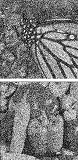}        &
	\hspace{-0.15in}\includegraphics[width=.068\textwidth]{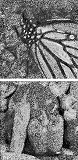}        &
	\hspace{-0.15in}\includegraphics[width=.092\textwidth]{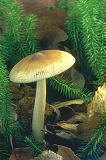}        & 
    \hspace{-0.15in}\includegraphics[width=.092\textwidth]{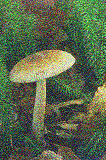}     &
	\hspace{-0.15in}\includegraphics[width=.092\textwidth]{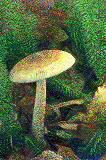} \\	
	\hspace{-0.10in}clean &\hspace{-0.15in} noisy &\hspace{-0.15in} mod & \hspace{-0.15in} clean &\hspace{-0.15in} org &\hspace{-0.15in} mod 
	
	
	
\end{tabular}
\vspace{-0.05in}
\caption{Our UPE improves the quality of the noisy input image. clean: clean image, noisy: noisy input image, mod: modified noisy image by UPE.
\label{fig:truncate_input_modify}}
\end{center}
\vspace{-0.25in}
\end{figure}

\begin{figure}[tb]
\begin{center}
    \begin{tabular}{c}
        \hspace{-0.10in}\includegraphics[width=.48\textwidth]{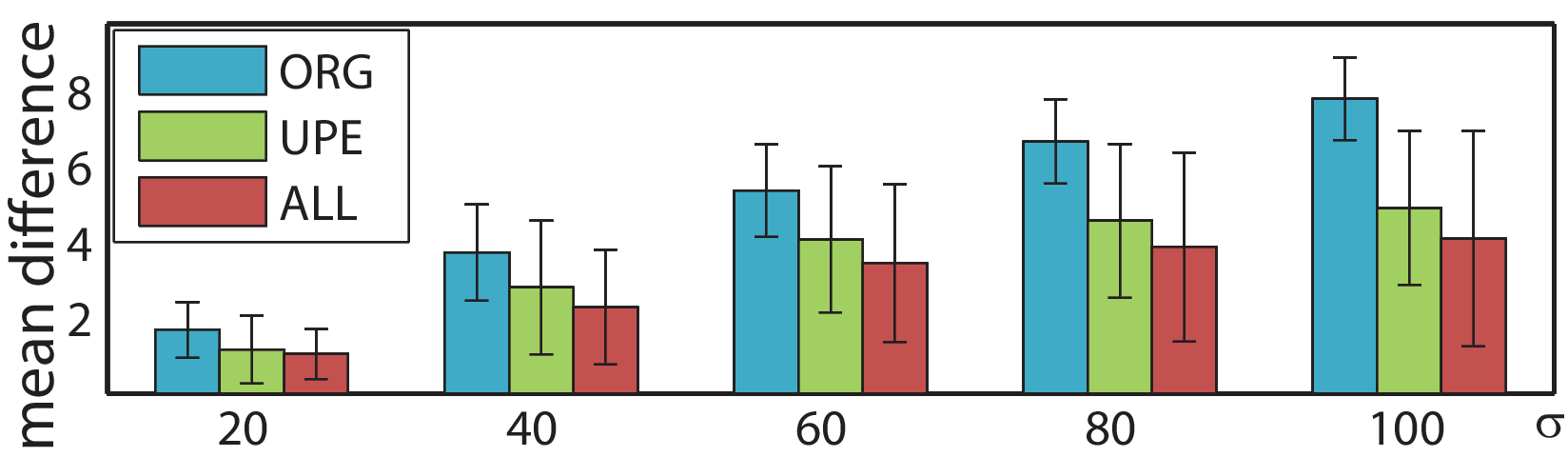}\\
	\end{tabular}
	\vspace{-0.05in}
	\caption{The difference between different groups of similar patches to the  ground truth reference patches. ORG: original groups. UPE: groups modified by UPE. ALL: groups modified by both UPE and clustering.\label{fig:upe01}}
\end{center}
\vspace{-0.20 in}
\end{figure}

Figure~\ref{fig:truncate_input_modify} shows that this simple and conservative strategy significantly improves the quality of the noisy input image. In addition, it can be seen that the UPE modifies the flat regions more. This is because initial patch searching  step  is  biased  more  in  flat  areas  than  textured  areas with strong structures. To statistically verify the effectiveness of UPE, we randomly sample 10,000 similar pixel groups and derive 3 versions of them: the original groups (ORG), the modified groups after UPE (UPE), and the final groups after both UPE and clustering (ALL), respectively. We report the average intensity difference between them and their corresponding ground truth references in Figure~\ref{fig:upe01}. It can be seen that UPE makes similar patches statistically closer to the ground truth than the original ones and clustering further reduces the difference.

There are two basic steps in a patch-based denoising method: finding a set of similar patches and recovering the clean patch from this patch set. The modified noisy input image can help 1) find a better set of patches and 2) better recover the clean patch since the unreliable pixels are corrected, in addition to having better similar patches.

\subsection{Complexity}
\label{sec:complexity}

Assume that each reference patch has $m$ $n$-dimensional similar patches in at most k clusters, the main computational cost of our algorithm is the unreliable pixel estimation and clustering. Since we use K-means++, the cost of the clustering is $O(mnk)$. The unreliable pixel estimation has $O(mn)$ operations. Suppose there are $l$ reference patches in the image, the total computational complexity is $O(lmnk)$. In our MATLAB implementation, given a 256$\times$256 input noisy image with standard deviation $\sigma=20$, the running time is about 10 seconds for our patch searching on BM3D, is about 10 seconds on  on a desktop with Intel(R) Core(TM) i7-4770 CPU (3.40GHz). While the running time of our patch searching part is similar on PLOW and LPCA, the actual speed is slower as extra overloaded is added to incorporate our patch searching into the authors' original code.

%% file: exp.tex
\newcolumntype{I}{!{\vrule width 2pt}}
\newlength\savedwidth
\newcommand\whline{\noalign{\global\savedwidth\arrayrulewidth
                            \global\arrayrulewidth 2pt}%
                   \hline
                   \noalign{\global\arrayrulewidth\savedwidth}}
\newlength\savewidth
\newcommand\shline{\noalign{\global\savewidth\arrayrulewidth
                            \global\arrayrulewidth 2pt}%
                   \hline
                   \noalign{\global\arrayrulewidth\savewidth}}

\subsection{Implementation Details}
\label{sec:details} 

We apply our algorithm for similar patch searching to three representative patch-based denoising methods, including BM3D~\cite{bm3d}, PLOW~\cite{plow} and LPCA~\cite{lpg-pca}. For BM3D, we use an open-source implementation (C++) proposed by Marc~\cite{bm3d_c}. For PLOW and LPCA, we utilize the source code from the author's website. All these methods share a common two-stage framework in which a preliminary denoised image is generated in the first stage to improve the similar patch searching in the second stage. Because similar patches with better similarities can benefit more from our clustering-based patch searching, we mainly apply our method to the similar patches obtained in the second denoising stage. We modify both the similar patches' intensities as well as the similar patch locations. The algorithms embedded with our method for better patch searching are denoted as BM3D$_{bst}$, PLOW$_{bst}$ and LPCA$_{bst}$, respectively. In addition, since LPCA uses 250 similar patches for image denoising, it may suffer from an over-smoothing problem. We thus modify LPCA by adaptively using NNS to search for the same number of similar patches as used in our LPCA$_{bst}$ for each noisy reference patch. We use this modified version of LPCA as an additional baseline for comparison and denote it as LPCA$_{bas}$.

We extend our modified denoising algorithms to denoise color images. Specifically, for PLOW and LPCA, we separately apply our boosted denoising algorithms on individual R,G,B channels, as proposed in the original algorithms. For BM3D, the noisy images are first transformed from RGB to YUV and the indices of similar patches are searched in the Y channel. The denoising is then separately performed on R,G,B channels using the same indices that have been obtained in the patch searching step. Thus, in BM3D$_{bst}$ for color image denoising, we apply our unreliable pixel estimation on all R,G,B and Y channels while the clustering-based patch searching is only performed on the Y channel. Finally, the R,G,B channels are denoised separately using our modified similar patches, in which the indices are searched in the Y channel and the pixel intensities are retrieved from the modified R,G,B channels, respectively.

\begin{figure}[b]
\vspace{-0.175in}
\begin{center}
    \begin{tabular}{cccccccc}
        \hspace{-0.10in}\includegraphics[width=.058\textwidth]{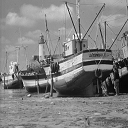}      &
		\hspace{-0.15in}\includegraphics[width=.058\textwidth]{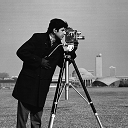} &
		\hspace{-0.15in}\includegraphics[width=.058\textwidth]{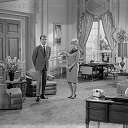}    &
		\hspace{-0.15in}\includegraphics[width=.058\textwidth]{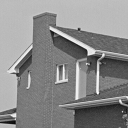}     &   

	    \hspace{-0.15in}\includegraphics[width=.058\textwidth]{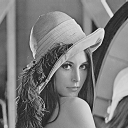}    &
		\hspace{-0.15in}\includegraphics[width=.058\textwidth]{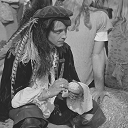}     &
		\hspace{-0.15in}\includegraphics[width=.058\textwidth]{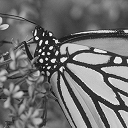} &
		\hspace{-0.15in}\includegraphics[width=.058\textwidth]{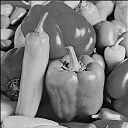} 
    \end{tabular}
	\vspace{-0.05in}	
    \caption{ Standard test images used in our experiments. \label{fig:images}}	
\end{center}
\vspace{-0.20in}
\end{figure}

\subsection{Parameter Settings}
\label{sec:para} 


For similar patch group denoising, we use exactly the same parameter settings as used in the original denoising methods. Different parameter settings are selected for the three denoising methods because 1) various quality of intermediate results are generated during denoising and 2) different denoising methodologies are exploited. 

Take the number of similar patch candidates $m$ as an example, denote $m_{org}$ as the number of similar patches used in the original denoising algorithms. For BM3D$_{bst}$ and PLOW$_{bst}$, as only a small amount of similar patch candidates are used in the original methods, we collect more initial similar patch candidates (100 and 30, respectively). While for LPCA, as 250 similar patches are used in the original method, suffering from an over-smoothing problem, we reduce $m$ to 150 in our boosted method. We choose a small number of patch candidates because 1) Nearest Neighbour Search already provides a reasonable set of similar patch candidates and 2) a large number of patch candidates are slow to process. 

Parameters $\alpha$ and $\beta$ in Equation~\ref{eq: grad_lambda} are selected according to how the image gradient is recovered in the basically denoised estimation. For BM3D, we set $\alpha = 2.5$ and $\beta=72$. We select larger $\alpha$ and $\beta$ values for PLOW (20,144) and LPCA (25,100) as their corresponding basic estimations are often over-smoothed, leading to biased gradient estimations with magnitudes smaller than the ground truth.

We select the maximal cluster number $K_{max}=4$ empirically. Specifically, we tested a large number of patches at a range of noise levels and found that for about 90\% of the patches, the optimal cluster numbers are less than or equal to 4. The maximal optimal cluster number is 6. More importantly, the denoising quality difference between 4 and 6 is negligible. Since a small $K_{max}$ value leads to a fast speed, we choose $K_{max}=4$ in all our experiments.

\vspace{-0.025in}
\subsection{Results on standard test images}
\label{sec:std_img} 
\vspace{-0.025in}


We test our method on 8 widely used standard test images as shown in Figure~\ref{fig:images}. To synthesize noisy images, we add White Gaussian noise with zero mean and standard deviation $\sigma=20,40,60,80,100$. Noisy pixels that are corrupted beyond the range of [0,255] are truncated. We use the PSNR to quantitatively evaluate the performance of our boosted patch-based denoising algorithms. In Table~\ref{tab:comp_03_state-of-art} we report the average PSNR results of the original similar patch-based denoising algorithms and our boosted algorithms. Higher PSNR results for each patch-based denoising algorithm on each noise level are highlighted in bold.

\begin{figure}[b]
\vspace{-0.25in}
  \begin{center}
    \begin{tabular}{c}
       \hspace{-0.125in}\includegraphics[width=.40\textwidth]{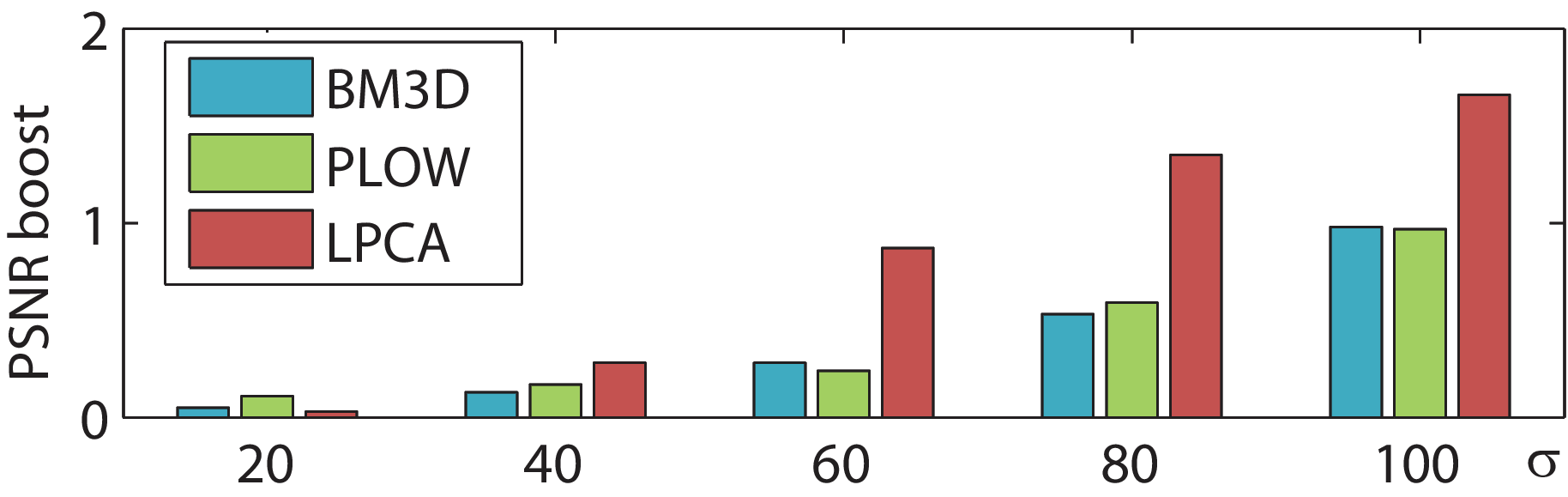} 
    \end{tabular}
	\vspace{-0.175in}
    \caption{The average PSNR improvement. \label{fig:psnr_bst}}  
  \end{center}
  \vspace{-0.22in}
\end{figure}

The results show that our similar patch searching algorithm can effectively improve the performance of the three patch-based denoising methods. By embedding our framework for similar patch searching, BM3D achieves an average improvement of 0.05dB--0.98dB across different noise levels. PLOW and LPCA's performance are improved by 0.11dB-0.97dB and 0.03dB--1.66dB, respectively. One can also see that while LPCA$_{bas}$ gains a little improvement over LPCA, our LPCA$_{bst}$ still significantly performs better. BM3D$_{bst}$ performs superior among the three boosted denoising algorithms. In addition, it can be seen in Figure~\ref{fig:psnr_bst} that images with higher noise levels gain larger improvements than the ones with lower noise levels. There are two reasons. First, heavy noise can bring a larger bias in the similar patch searching process. Second, images with higher level of noise have more unreliable pixels. Thus, these images can benefit more from our proposed similar patch search approach.

In Figure~\ref{fig:visual_01} we compare the visual performance of BM3D, PLOW and LPCA with their improved versions. It can be seen that the improved algorithms that incorporated with our patch searching method better recover fine image details as well as better preserve color fidelity.

\begin{figure}[b]
\vspace{-0.20in}
  {\footnotesize
  \begin{center}
      \begin{tabular}{cccc}
        \hspace{-0.05in}\includegraphics[width=.11\textwidth]{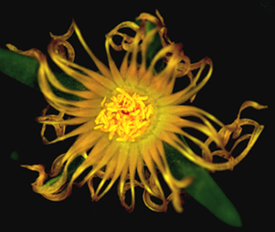} & 
		\hspace{-0.14in}\includegraphics[width=.11\textwidth]{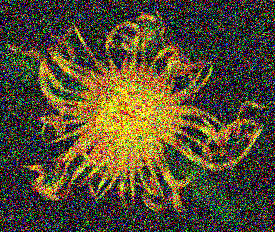} &
		\hspace{-0.14in}\includegraphics[width=.11\textwidth]{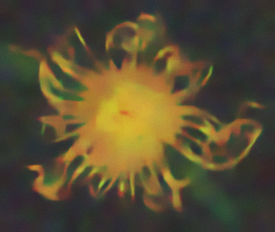} &
		\hspace{-0.14in}\includegraphics[width=.11\textwidth]{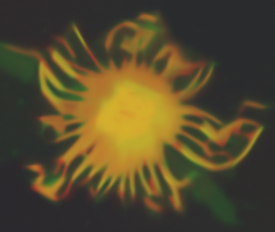}\\	
	    \hspace{-0.05in}(a) Clean image & \hspace{-0.14in}(b) Noisy image   & \hspace{-0.14in}(c) EPLL    & \hspace{-0.14in}(d) WNNM\\	
		\hspace{-0.05in}                & \hspace{-0.14in} $\sigma = 100$   & \hspace{-0.14in} PSNR=20.48 & \hspace{-0.14in}PSNR=18.26\\	
		
	    \hspace{-0.05in}\includegraphics[width=.11\textwidth]{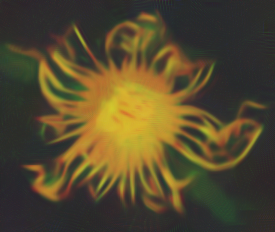} & 
		\hspace{-0.14in}\includegraphics[width=.11\textwidth]{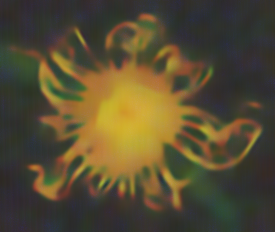} &
		\hspace{-0.14in}\includegraphics[width=.11\textwidth]{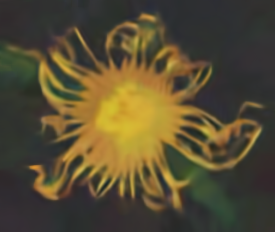} & 
		\hspace{-0.14in}\includegraphics[width=.11\textwidth]{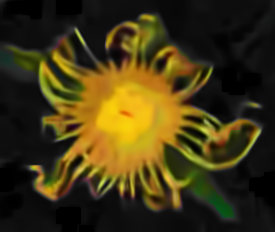}\\	
		\hspace{-0.05in}(e) PGPD    & \hspace{-0.14in}(f) PCLR    & \hspace{-0.14in}(g) BM3D       & \hspace{-0.14in}(h) BM3D$_{bst}$\\	
		\hspace{-0.05in} PSNR=18.57 & \hspace{-0.14in} PSNR=17.32 & \hspace{-0.14in}    PSNR=17.97 & \hspace{-0.14in}PSNR=23.13\\

      \end{tabular}
	  \vspace{-0.10in}
      \caption{Comparison of BM3D$_{bst}$ and other methods.\label{fig:firsteg}}
	  \vspace{0.07in}

      \begin{tabular}{cccc}
        \hspace{-0.05in}\includegraphics[width=.11\textwidth]{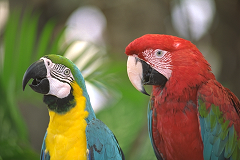} & 
		\hspace{-0.14in}\includegraphics[width=.11\textwidth]{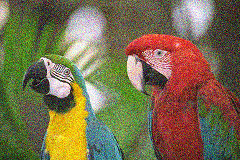} & 
		\hspace{-0.14in}\includegraphics[width=.11\textwidth]{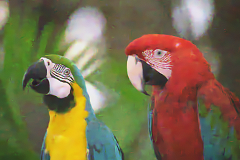} &  
		\hspace{-0.14in}\includegraphics[width=.11\textwidth]{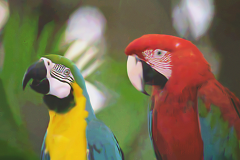}\\	
		\hspace{-0.05in}(a) Clean image & \hspace{-0.14in}(b) Noisy image   & \hspace{-0.14in}(c) EPLL    & \hspace{-0.14in}(d) WNNM\\	
		\hspace{-0.05in}                & \hspace{-0.14in} $\sigma = 40$   & \hspace{-0.14in} PSNR=29.22 & \hspace{-0.14in}PSNR=29.53\\

	    \hspace{-0.05in}\includegraphics[width=.11\textwidth]{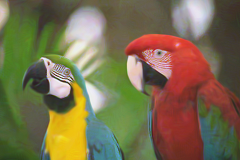} &
		\hspace{-0.14in}\includegraphics[width=.11\textwidth]{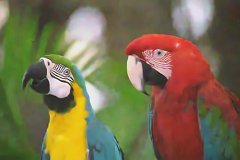} & 
		\hspace{-0.14in}\includegraphics[width=.11\textwidth]{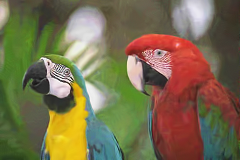} &
		\hspace{-0.14in}\includegraphics[width=.11\textwidth]{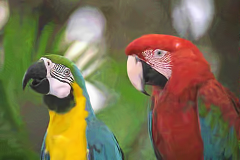}\\	
		\hspace{-0.05in}(e) PGPD    & \hspace{-0.14in}(f) PCLR    & \hspace{-0.14in}(g) BM3D       & \hspace{-0.14in}(h) BM3D$_{bst}$\\	
		\hspace{-0.05in} PSNR=29.07 & \hspace{-0.14in} PSNR=29.11 & \hspace{-0.14in}    PSNR=30.31 & \hspace{-0.14in}PSNR=30.83\\
		
      \end{tabular}
	  \vspace{-0.10in}
      \caption{Comparison of BM3D$_{bst}$ and other methods.\label{fig:secondeg}}	
	  \vspace{0.07in}

      \begin{tabular}{cccc}
        \hspace{-0.05in}\includegraphics[width=.11\textwidth]{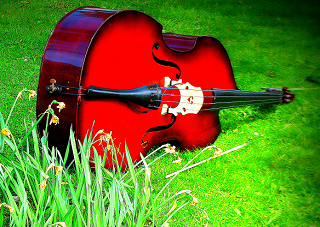} & 
		\hspace{-0.14in}\includegraphics[width=.11\textwidth]{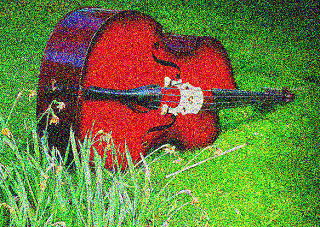} & 
		\hspace{-0.14in}\includegraphics[width=.11\textwidth]{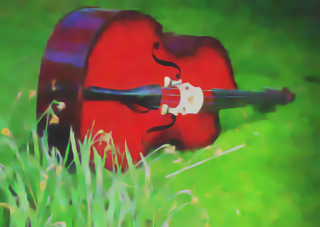} &  
		\hspace{-0.14in}\includegraphics[width=.11\textwidth]{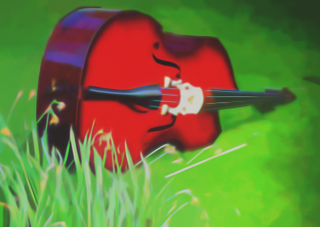}\\	
		\hspace{-0.05in}(a) Clean image & \hspace{-0.14in}(b) Noisy image   & \hspace{-0.14in}(c) EPLL    & \hspace{-0.14in}(d) WNNM\\	
		\hspace{-0.05in}                & \hspace{-0.14in} $\sigma = 100$   & \hspace{-0.14in} PSNR=15.59 & \hspace{-0.14in}PSNR=15.64\\	
		
	
	    \hspace{-0.05in}\includegraphics[width=.11\textwidth]{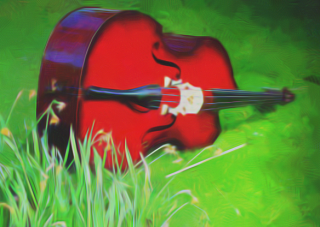} &
		\hspace{-0.14in}\includegraphics[width=.11\textwidth]{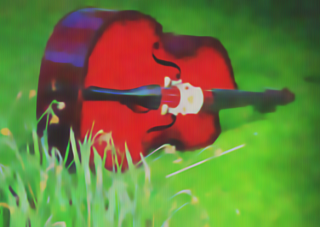} & 
		\hspace{-0.14in}\includegraphics[width=.11\textwidth]{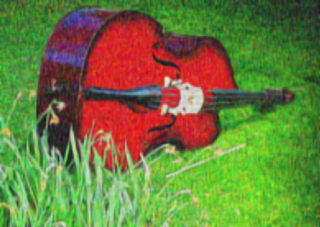} &
		\hspace{-0.14in}\includegraphics[width=.11\textwidth]{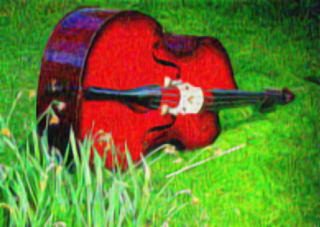}\\	
		\hspace{-0.05in}(e) PGPD    & \hspace{-0.14in}(f) PCLR    & \hspace{-0.14in}(g) PLOW       & \hspace{-0.14in}(h) PLOW$_{bst}$\\	
		\hspace{-0.05in} PSNR=15.95 & \hspace{-0.14in} PSNR=15.10 & \hspace{-0.14in}    PSNR=15.36 & \hspace{-0.14in}PSNR=16.74\\
		
      \end{tabular}
	  \vspace{-0.10in}
      \caption{Comparison of PLOW$_{bst}$ and other methods.\label{fig:thirdeg}}
      \vspace{0.05in}
      	  
  \end{center}}
  \vspace{-0.15in}
\end{figure}

We also compare BM3D$_{bst}$, which performs superior among the three improved algorithms, with the recent state-of-the-art denoising algorithms, including EPLL~\cite{epll}, WNNM~\cite{wnnm}, PGPD~\cite{xu:pgpd} and PCLR~\cite{chen:external} in Table~\ref{tab:comp_03_state-of-art}. It can be seen that BM3D$_{bst}$ generally outperforms all other methods on images with noise levels higher than $\sigma=60$. For images with $\sigma=20$ and $40$, BM3D$_{bst}$ has comparable performance with other methods. The visual comparisons of these competing denoising methods on images with heavy noise are shown in Figure~\ref{fig:firsteg},~\ref{fig:secondeg} and~\ref{fig:thirdeg}. Our boosted methods significantly outperform the other methods by as much as 1.88dB, 1.99dB and 0.79dB, respectively. Fine image structures as well as the color fidelity are therefore better reconstructed by our boosted denoising methods.  

\begin{table} [htb]
  {\scriptsize
  \begin{center}
    \caption{Average PSNR results of the competing methods\label{tab:comp_03_state-of-art}}
    \vspace{-0.10in}
	\begin{tabular}{p{12mm}p{8mm}p{8mm}p{8mm}p{8mm}p{8mm}}  	
	  \hline
	  $\sigma$           &       20       &       40       &       60       &       80       &       100      \\
	  \hline 
	  BM3D              &      31.34     &      27.97     &      25.71     &      23.82     &      22.10     \\
	  BM3D$_{bst}$      & \textbf{31.39} & \textbf{28.11} & \textbf{26.00} & \textbf{24.35} & \textbf{23.08} \\
	  PLOW              &      30.34     &      27.55     &      25.38     &      23.51     &      21.82     \\
	  PLOW$_{bst}$      & \textbf{30.46} & \textbf{27.72} & \textbf{25.62} & \textbf{24.10} & \textbf{22.79} \\
	  LPCA              &      30.69     &      26.89     &      24.31     &      22.51     &      21.10     \\
	  LPCA$_{bas}$      &      30.69     &      26.91     &      24.36     &      22.58     &      21.19     \\
	  LPCA$_{bst}$      & \textbf{30.72} & \textbf{27.16} & \textbf{25.18} & \textbf{23.86} & \textbf{22.76} \\
	  \hline
	  EPLL              &      31.16     &      27.83     &      25.41     &      23.13     &      21.12     \\
	  PGPD              &      31.38     &      28.04     &      25.59     &      23.63     &      21.92     \\
	  PCLR              &      31.58     &      28.11     &      25.46     &      23.03     &      20.91     \\
	  WNNM              & \textbf{31.61} & \textbf{28.12} &      25.50     &      23.28     &      21.37     \\
	  \hline  
    \end{tabular}
  \end{center}}
  \vspace{-0.2in}
\end{table}	

\begin{table} [htb]
  {\scriptsize
  \begin{center}
      \caption{Average PSNR results on different levels of Poisson noise \label{tab:comp_02_pos}}
      \vspace{-0.20in}
	  \begin{tabular}{p{12mm}p{8mm}p{8mm}p{8mm}p{8mm}p{8mm}}	  
	    \hline
	    $\kappa$            &       20       &       35       &       50       &       65       &       80       \\
	    \hline
	    BM3D              &      26.92     &      25.22     &      24.03     &      22.83     &      21.93     \\
	    BM3D$_{bst}$      & \textbf{27.12} & \textbf{25.65} & \textbf{24.52} & \textbf{23.55} & \textbf{22.31} \\
	    PLOW              &      25.35     &      22.90     &      22.17     &      21.79     &      20.80     \\
	    PLOW$_{bst}$      & \textbf{25.48} & \textbf{23.31} & \textbf{22.62} & \textbf{21.98} & \textbf{21.06} \\
	    LPCA              &      25.64     &      23.67     &      22.59     &      21.29     &      20.71     \\
	    LPCA$_{bas}$      &      25.65     &      23.74     &      22.52     &      21.39     &      20.59     \\
	    LPCA$_{bst}$      & \textbf{26.04} & \textbf{24.34} & \textbf{23.19} & \textbf{21.57} & \textbf{21.22} \\
	    \hline  
      \end{tabular}
  \end{center}}
  \vspace{-0.25in}
\end{table}

\begin{table} [htb]
  {\scriptsize
  \vspace{-0.05in}
  \begin{center}
	\caption{Average PSNR results of BM3D$_{bst}$ and POD\label{tab:patch_ordering}}
	\vspace{-0.10in}
	\begin{tabular}{p{35mm}p{35mm}}
	  \hline
	  POD ($\sigma=20/50/75$)         & BM3D$_{bst}$ ($\sigma=20/50/75$)  \\	
	  $31.04/26.70/24.19$             & $\textbf{31.39}/\textbf{27.03}/\textbf{24.75}$\\
	  \hline
    \end{tabular}
  \end{center}}
  \vspace{-0.25in}
\end{table}		

\begin{table} [htb]
  {\scriptsize
  \vspace{-0.05in}
  \begin{center}
	\caption{Comparison of clustering methods on BM3D$_{bst}$ \label{tab:comp_03_dif_cluster}}
	\vspace{-0.10in}
	\begin{tabular}{p{15mm}p{8mm}p{8mm}p{8mm}p{8mm}p{8mm}} 		  
	  \hline
	  $\sigma$          &   20  &   40  &   60  &   80  &  100  \\
	  \hline
	  K-means++         & 31.39 & 28.11 & 26.00 & 24.35 & 23.08 \\
	  GMM               & 31.37 & 28.08 & 25.95 & 24.31 & 23.02 \\
	  LSC               & 31.37 & 28.04 & 25.91 & 24.23 & 22.92 \\
	  \hline  
    \end{tabular}
  \end{center}}
\vspace{-0.25in}	
\end{table}

\begin{table} [htb]
  {\scriptsize
  \begin{center}
    \caption{Average SSIM scores of the competing methods\label{tab:ssim_comp}}
    \vspace{-0.10in}
	\begin{tabular}{p{12mm}p{8mm}p{8mm}p{8mm}p{8mm}p{8mm}}  	
	  \hline
	  $\sigma$           &       20       &       40       &       60       &       80       &       100      \\
	  \hline 
	  BM3D              &      .8655     &      .7867     &      .7285     &      .6823     &      .6401     \\
	  BM3D$_{bst}$      & \textbf{.8664} & \textbf{.7897} & \textbf{.7342} & \textbf{.6889} & \textbf{.6486} \\
	  \hline
	  EPLL              &      .8629     &      .7805     &      .7124     &      .6591     &      .6027     \\
	  PGPD              &      .8606     &      .7863     &      .7231     &      .6687     &      .6219     \\
	  PCLR              &      .8668     & \textbf{.7898} &      .7233     &      .6563     &      .5955     \\
	  WNNM              & \textbf{.8672} &      .7891     &      .7238     &      .6646     &      .6122     \\
	  \hline  
    \end{tabular}
  \end{center}}
  \vspace{-0.25in}
\end{table}	

\begin{table} [htb]
  {\scriptsize
  \vspace{-0.04in}
  \begin{center}	
    \caption{Comparison on the BSD test images\label{tab:bsd}}
    \vspace{-0.10in}
	\begin{tabular}{p{8mm}p{8mm}p{8mm}p{8mm}p{8mm}p{8mm}} 
	  \hline
	  $\sigma$          &      20        &      40        &      60        &      80        &      100       \\
	  \hline
	\end{tabular}
	\begin{tabular}{p{8mm}p{8mm}p{8mm}p{8mm}p{8mm}p{8mm}}  
	  BM3D              &      29.35     &      25.89     &      23.86     &      22.22     &      20.79     \\
	  BM3D$_{bst}$      &\textbf{29.45}  & \textbf{26.16} & \textbf{24.35} & \textbf{23.04} & \textbf{22.04} \\
	  PLOW              &      27.58     &      25.26     &      23.51     &      21.78     &      20.32     \\
	  PLOW$_{bst}$      &\textbf{27.66}  & \textbf{25.43} & \textbf{23.96} & \textbf{22.68} & \textbf{21.62} \\
	  LPCA              &      28.98     &      25.40     &      23.34     &      21.91     &      20.81     \\
	  LPCA$_{bas}$      &      29.00     &      25.43     &      23.40     &      21.96     &      20.85     \\
	  LPCA$_{bst}$      &\textbf{29.07}  & \textbf{25.77} & \textbf{24.19} & \textbf{23.18} &\textbf{22.33}  \\
	  \hline
	  PGPD              &      29.30     &      25.82     &      23.71     &      22.13     &      20.79     \\
	  PCLR              &\textbf{29.55}  &      25.87     &      23.50     &      21.51     &      19.91     \\
	  EPLL              &      29.45     &      26.01     &      23.87     &      22.03     &      20.65     \\
	  WNNM              &      29.46     &      25.88     &      23.59     &      21.79     &      20.38     \\	 
	  \hline
    \end{tabular}
  \end{center}}
  \vspace{-0.25in}
\end{table}	

To test the robustness of our method across different noise types, we apply the three boosted denoising algorithms to images corrupted by Poisson noise. To add Poisson noise to a clean image $I$, we use $N=\kappa\cdot$poissrnd$(I/\kappa)$ in MATLAB as proposed by Zhang et al.~\cite{mult-v-denoise}, where poissrnd$(x)$ generates a Poisson distributed random number with mean and variance of $x$. We set $\kappa=20,35,50,65$ and $80$ to add different levels of Poisson noise. We report the average denoising performance in Table~\ref{tab:comp_02_pos}. One can see that the three patch-based methods still benefit from our patch searching method on images with Poisson noise.

\begin{figure}[t]
\centering
\vspace{-0.05 in}
    \includegraphics[width=.40\textwidth]{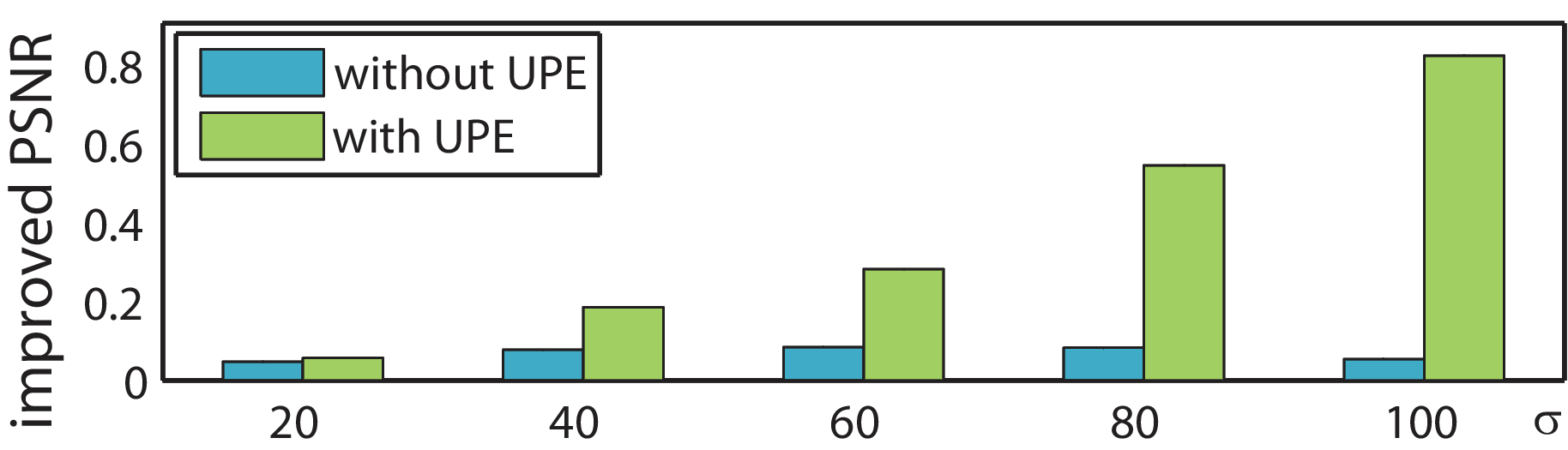}\\
	\vspace{-0.05in}
    \caption{The effect of our method with and without UPE. \label{fig:leaveUPEout}}
\vspace{-0.25 in}
\end{figure}

To evaluate the effect of unreliable pixel estimation (UPE), we leave UPE out and compare the corresponding PSNR improvement to that of our method with UPE in Figure~\ref{fig:leaveUPEout}. It shows that UPE contributes significantly to the final result, especially with high levels of noise, and clustering further boosts the denoising performance.

We also compare our BM3D$_{bst}$ to the Patch Ordering (POD)~\cite{Patch_ordering} method that selects better sets of similar patches for image denoising via patch ordering. We compare these two methods on images with noise levels $\sigma=20,50$ and $75$, as the parameter settings for those noise levels in the Patch Ordering method have been optimally set in the authors' code. We report the average PSNR results in Table~\ref{tab:patch_ordering}. It can be seen that our method outperforms POD.

We test our similar patch searching algorithm using different clustering methods on BM3D$_{bst}$, including Gaussian Mixture Model~\cite{mdlcluster} clustering and Landmark-based Spectral Clustering~\cite{lsc}, and compare them with K-means++ in Table~\ref{tab:comp_03_dif_cluster}. The results show that all three clustering algorithms have comparable performance. 

We report SSIM scores of our boosted BM3D$_{bst}$ in Table~\ref{tab:ssim_comp}, which shows our method improves BM3D consistently. Our boosted PLOW$_{bst}$ and LPCA$_{bst}$ have similar performance improvement. We skip their quantitative results due to the lack of space.

\vspace{-0.05in}
\subsection{Results on the BSD dataset}
\label{sec:bsd_img}
\vspace{-0.05in}


We use the 100 test images from the Berkeley Segmentation Dataset~\cite{bsd_dataset} to generate 500 noisy images with 5 different noise levels and test our method on them. In Table~\ref{tab:bsd} we compare our boosted denoising methods with the original patch-based denoising algorithms as well as the competing methods. One can see that the boosted algorithms robustly outperform the original ones. Statistically, among all the 500 noisy images, the improved denoising methods, including BM3D$_{bst}$, PLOW$_{bst}$ and LPCA$_{bst}$, beat their corresponding original methods for 482, 397 and 499 times, respectively. In addition, comparing with other state-of-the-art denoising methods, BM3D$_{bst}$ has comparable performance on images with $\sigma=20$ and performs better on higher noise levels. More visual examples of the boosted algorithms can be find in the supplementary material.

